\documentclass[conference]{IEEEtran}
\IEEEoverridecommandlockouts

\usepackage{cite}
\usepackage{amsmath,amssymb,amsfonts}
\usepackage{graphicx}
\usepackage{textcomp}
\usepackage{xcolor}

\usepackage{tikz}

\usepackage{mathtools}
\usepackage{booktabs}
\usepackage{algorithm} 
\usepackage{algpseudocode}

\def\BibTeX{{\rm B\kern-.05em{\sc i\kern-.025em b}\kern-.08em
    T\kern-.1667em\lower.7ex\hbox{E}\kern-.125emX}}
    
\DeclareMathOperator*{\argmaxA}{arg\,max}

\algnewcommand{\algorithmicand}{\textbf{ AND }}
\algnewcommand{\AND}{\algorithmicand}

\newcommand\copyrighttext{%
  \footnotesize \textcopyright 2020 IEEE. Personal use of this material is permitted. Permission from IEEE must be obtained for all other uses, in any current or future media, including reprinting/republishing this material for advertising or promotional purposes, creating new collective works, for resale or redistribution to servers or lists, or reuse of any copyrighted component of this work in other works. Accepted to be Published in: Proceedings of the IJCNN 2020: International Joint Conference on Neural Networks, 19 - 24 July, 2020, Glasgow (UK)}
\newcommand\copyrightnotice{%
\begin{tikzpicture}[remember picture,overlay]
\node[anchor=south,yshift=10pt] at (current page.south) {\fbox{\parbox{\dimexpr\textwidth-\fboxsep-\fboxrule\relax}{\copyrighttext}}};
\end{tikzpicture}%
}  

\begin{document}

\title{Seasonal Averaged One-Dependence Estimators: A Novel Algorithm to Address Seasonal Concept Drift in High-Dimensional Stream Classification}

\author{\IEEEauthorblockN{Rakshitha Godahewa\IEEEauthorrefmark{1},
Trevor Yann\IEEEauthorrefmark{2}, 
Christoph Bergmeir\IEEEauthorrefmark{1},
Francois Petitjean\IEEEauthorrefmark{1}}
\IEEEauthorblockA{\IEEEauthorrefmark{1}Faculty of Information Technology\\
Monash University, Melbourne, Australia\\
rakshitha.godahewa@monash.edu, christoph.bergmeir@monash.edu, francois.petitjean@monash.edu}
\IEEEauthorblockA{\IEEEauthorrefmark{2}Seek Group\\
Melbourne, Australia\\
trevor.yann@gmail.com}
}

\maketitle

\copyrightnotice

\begin{abstract}
Stream classification methods classify a continuous stream of data as new labelled samples arrive. They often also have to deal with concept drift. This paper focuses on seasonal drift in stream classification, which can be found in many real-world application data sources. Traditional approaches of stream classification consider seasonal drift by including seasonal dummy/indicator variables or building separate models for each season. But these approaches have strong limitations in high-dimensional classification problems, or with complex seasonal patterns. This paper explores how to best handle seasonal drift in the specific context of news article categorization (or classification/tagging), where seasonal drift is overwhelmingly the main type of drift present in the data, and for which the data are high-dimensional. We introduce a novel classifier named Seasonal Averaged One-Dependence Estimators (SAODE), which extends the AODE classifier to handle seasonal drift by including time as a super parent. We assess our SAODE model using two large real-world text mining related datasets each comprising approximately a million records, against nine state-of-the-art stream and concept drift classification models, with and without seasonal indicators and with separate models built for each season. Across five different evaluation techniques, we show that our model consistently outperforms other methods by a large margin where the results are statistically significant.
\end{abstract}


\section{Introduction}
\label{sec:intro}

Stream classifiers are able to learn a model and refine it as more labelled data progressively are available\cite{ref_41}, providing scalability and responsiveness for classification of large data streams. Many real-world applications involve stream classification including text classification, power load classification, network traffic analysis and high volume social network feeds, making it an emerging area of research in the fields of data mining, knowledge discovery and machine learning.

\begin{figure}[t]
\includegraphics[width=\columnwidth]{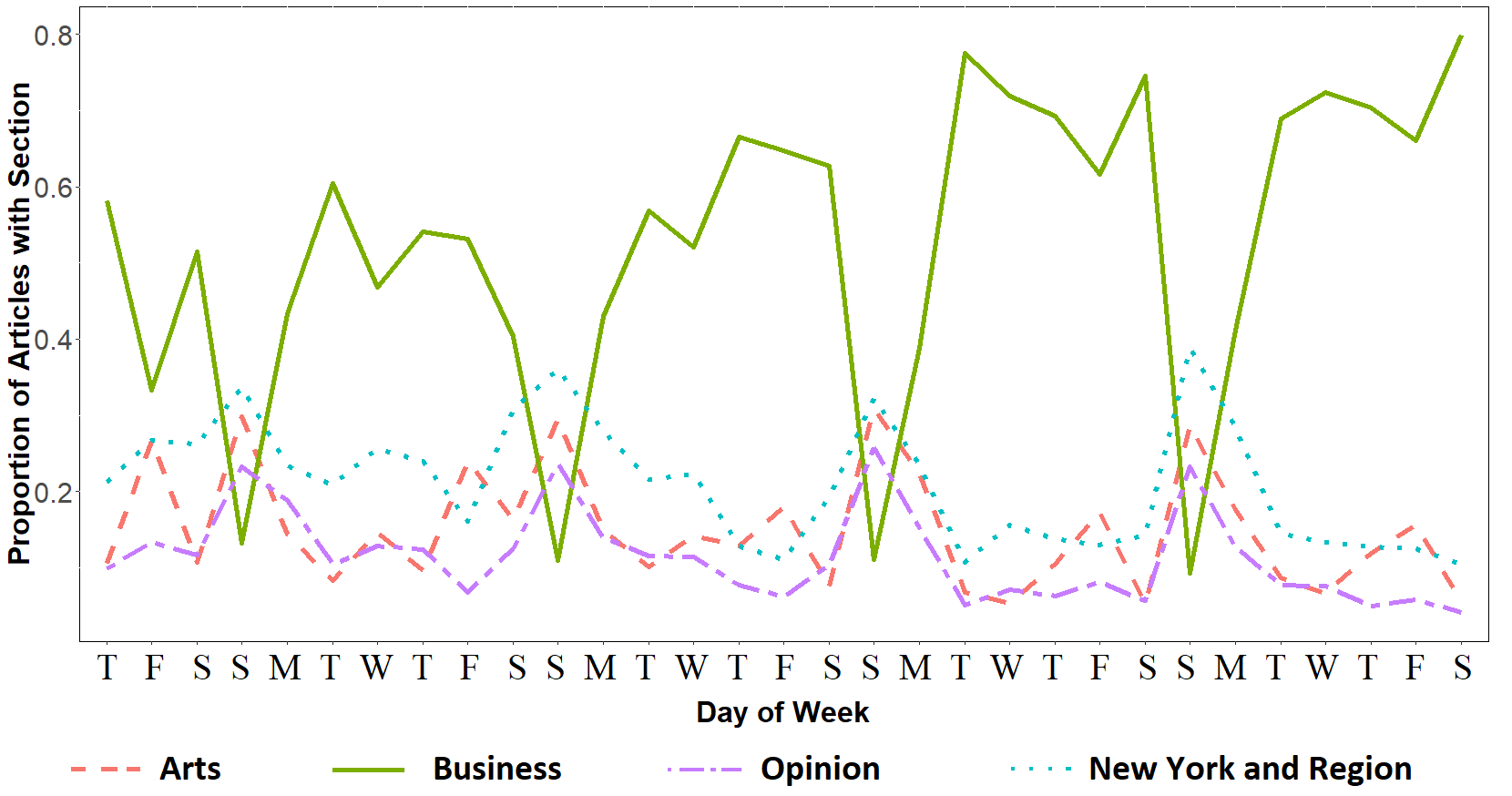}
\caption{Article proportions for the top four online sections of the NYT dataset: \textit{Arts}, \textit{Business}, \textit{Opinion} and \textit{New York and Region} over article day of the week for the first 31 days of the dataset. We note a strong weekly seasonality in the proportions of all online sections.}
\label{fig3}
\end{figure}

A variety of stream classification algorithms have been proposed over past years \cite{ref_35,ref_34,ref_40,ref_33,ref_36,ref_37}. Some of these classifiers \cite{ref_33, ref_36, ref_37} are concept drift classifiers as they can handle concept drift issues including the changes in data interpretations, class proportions and input-output relationships that can degrade the classifier accuracy over time \cite{ref_20}. However, most of them do not consider seasonality effects in designing the classification model. Many real-world datasets show strong seasonal patterns as shown in Fig.~\ref{fig3}. Here, we can see that in a dataset of New York Times (NYT) articles, there is a strong seasonal pattern in the sense that there are more Arts, Opinion and New York and Region articles on Sundays than during the other days, with a decrease of Business articles. This type of phenomenon leads to two potential issues:
\begin{enumerate}
    \item Classification accuracy will be significantly impacted when using a classifier that does not handle drift, and only has the capacity to be incrementally updated, such as a Hoeffding tree \cite{ref_35}, and does not know the article date. We show simply providing the day of the week as a separate variable only mitigates the problem in our experiments. 
    \item Using a stream classifier that can handle drift in the distribution will have to forget a lot of what it has learnt at the start of each period. This is because most drift classifiers are built to forget part of the past every time a change in the distribution is uncovered. This reason has been hypothesised before as a reason for decreasing accuracy of classifiers over time \cite{ref_20}. We clearly observe it in our experiments (Section~\ref{sec:exp}) where drift classifiers actually perform worse than algorithms not built to handle any sort of drift even if provided with a variable indicating the period of the week. 
\end{enumerate}

\begin{figure}[t]
\includegraphics[width=\columnwidth]{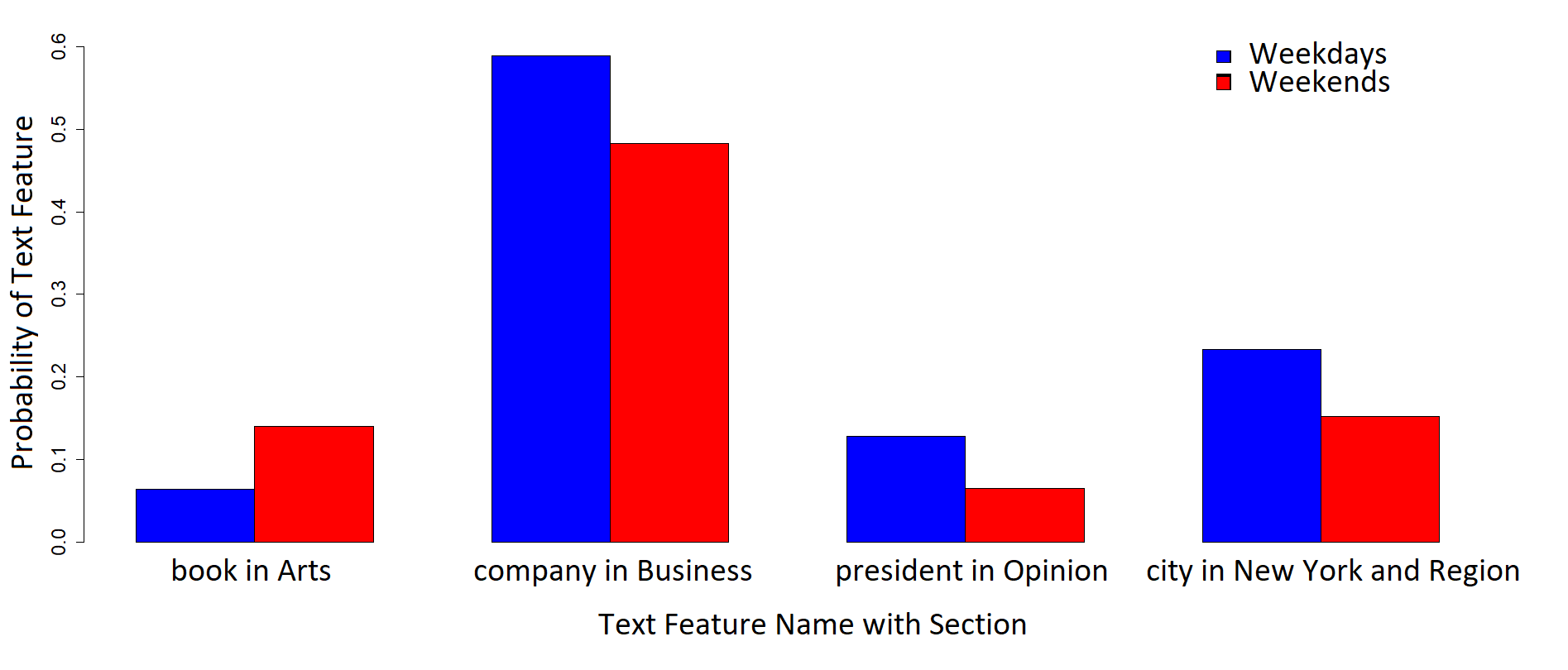}
\caption{Probabilities of appearing four text features: ``book", ``company", ``president" and ``city" in the top four online sections of the NYT Dataset: \textit{Arts}, \textit{Business}, \textit{Opinion} and \textit{New York and Region}, respectively, in weekdays and weekends for the entire dataset. The graph shows the variation of feature impact on stream classification in weekdays and weekends. It indicates seasonal concept drift may occur in the dataset.}
\label{fig10}
\end{figure}

In addition, seasonal changes are naturally not limited to changes in the class distribution but will often impact other aspects of the data distribution. This is illustrated in Fig.~\ref{fig10} where we can see that in the category/class of articles concerning the Arts, the term `book' occurs more often during the weekend edition than during the week; a pattern due to the `New York Times Sunday Book Review' section. 

A simple way to handle seasonality is to include an indicator variable (e.g. `day of the week') to the data. However, as we show in Section~\ref{sec:exp}, for real-world applications that are high-dimensional, that variable ends up being lost among thousands of other variables, with state-of-the-art classifiers failing to be beneficial. 

Training multiple classifiers, one per each season, is another possible approach to handle seasonality. However, we show in Section~\ref{sec:exp} that it is not the best mechanism to handle seasonality in high-dimensional stream classification.

    
Averaged One-Dependence Estimators (AODE) are an improved version of Na\"ive Bayes (NB) that relax the attribute independence assumption \cite{ref_2} of NB. AODE is a fast, updatable and accurate classifier that facilitates including additional variables during the classification process.

In this paper, we make the following contributions:
\begin{enumerate}
\item We propose a novel classification model specifically designed to tackle high-dimensional stream classification with seasonal concept drift. Our model works with discrete attributes and leverages the AODE algorithm to be robust to seasonal variations while not forgetting about any past data. For this reason, we call our model Seasonal Averaged One-Dependence Estimators (SAODE).
\item We present a quantitative comparison of the relative performance of our SAODE model against nine state-of-the-art classifiers including both stream and concept drift classifiers, and using two real-world text-mining related datasets: the improved version of Reuters Corpus Volume 1 dataset or simply RCV1-v2 dataset \cite{ref_1} and the NYT dataset, each with over 800,000 instances. Our model outperforms all standard stream and concept drift classifiers used in the experiment by 10\% accuracy where the results are statistically significant. 
\item Finally, all SAODE related implementations, experiments and preprocessed datasets are publicly available at: https://github.com/rakshitha123/SAODE
\end{enumerate}

\begin{figure}[t]
\includegraphics[width=\columnwidth]{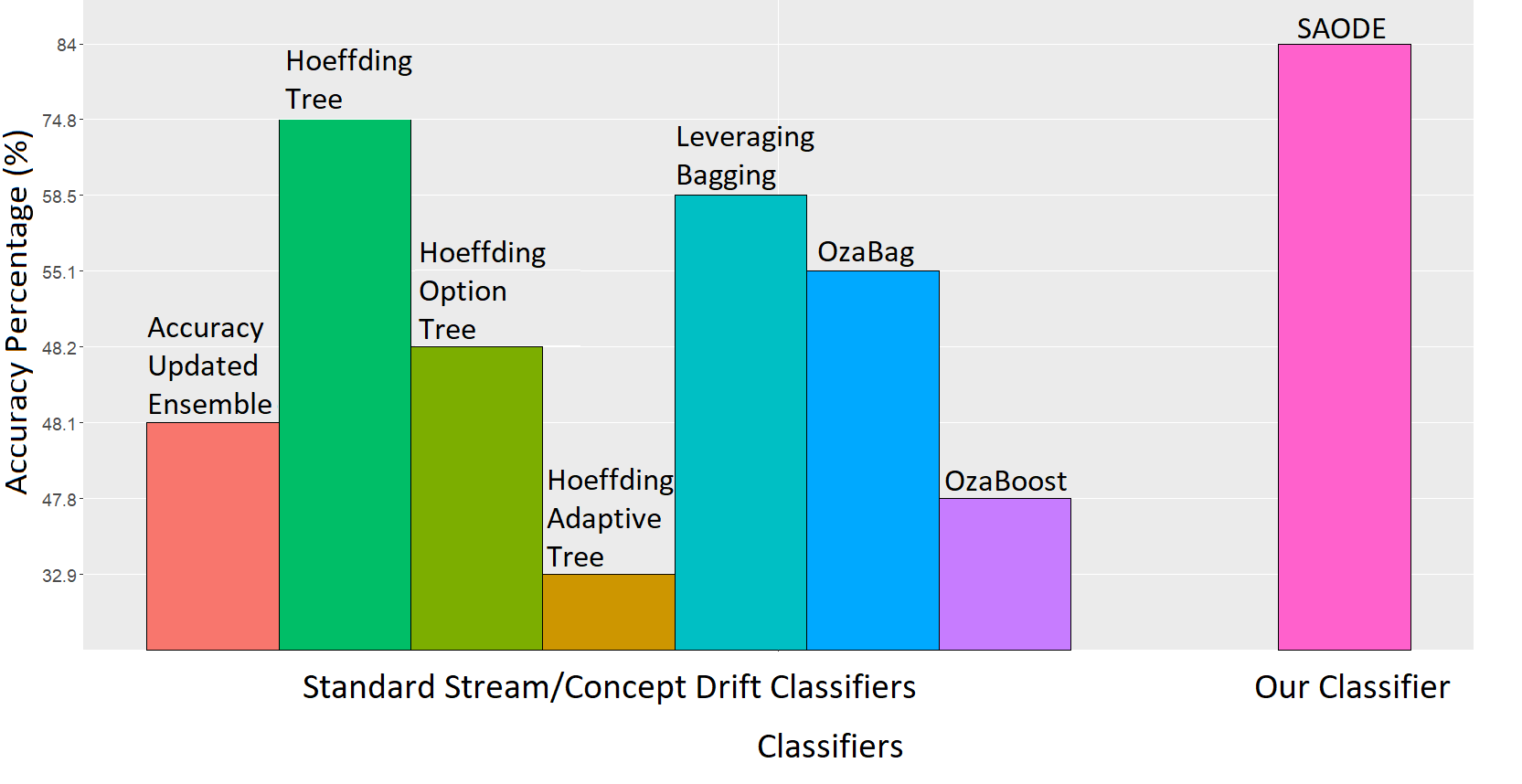}
\caption{Classification accuracy percentages calculated for the RCV1-v2 dataset in news stories classification for the proposed classifier: SAODE and the standard stream and concept drift classifiers: Accuracy Updated Ensemble, Hoeffding Tree, Hoeffding Option Tree, Hoeffding Adaptive Tree, Leveraging Bagging, OzaBag and OzaBoost. SAODE demonstrates a classification accuracy of 84.0\% and outperforms all other standard stream/concept drift classifiers by 10\% accuracy.} 
\label{fig11}
\end{figure}

The remainder of this paper is organized as follows: Section~\ref{sec:relw} reviews some related work. Section~\ref{sec:framework} introduces the basic concept of AODE. We present our main contributions in Section~\ref{sec:saode} and Section~\ref{sec:exp}. Section~\ref{sec:conclusion} concludes the paper.

\section{Related Work}
\label{sec:relw}

In this section, we review existing methods that are relevant to address stream classification and concept drift. 

A stream classifier must start providing predictions when only a subset of the data is available. It can assume either that the samples are independent and identically distributed or that the data distribution changes. The Hoeffding Tree \cite{ref_35} is a stream classifier that captures the data distribution more completely as data are made available. The Hoeffding Option Tree \cite{ref_34} is a derivative of the Hoeffding Tree that extends trees by maintaining more splits at each node acting as a type of hybrid ensemble within the tree. OzaBag and OzaBoost~\cite{ref_40} adapt bagging and boosting respectively to a stream classification context. 

Stream classification can also target problems where there is concept drift.
There are different forms of concept drift that can affect the effectiveness of machine learning models including virtual concept drift and real concept drift \cite{ref_19}. A quantitative distinction between different types of concept drift is given by Webb et al.~\cite{ref_20}. Multi-stream classification approaches also use techniques for drift detection \cite{ref_54}. Masud et al.~\cite{ref_55} present a new class detection method that can be integrated with traditional classifiers in the presence of concept drift. 

Concept drift adaption is required to overcome the resulting problems of concept drift \cite{ref_19}. The Concept-adapting Very Fast Decision Tree (CVFDT) \cite{ref_32} updates tree statistics as new examples are processed. Hoeffding Adaptive Trees \cite{ref_33} are an extension of Hoeffding Trees that modify the tree branching strategy to address concept drift. They use ADWIN method \cite{ref_33} to detect drift by considering all possible large subwindows for a distinct enough change where statistical tests are used to determine sufficiency in the subwindow sizes. Ensemble classifiers including the Accuracy Updated Ensemble \cite{ref_36}, Leveraging Bagging \cite{ref_37} and Dynamic Weighted Majority \cite{ref_57}, and non-ensemble classifiers including Support Vector Machines (SVM) \cite{ref_38} and k-Nearest Neighbour (k-NN) \cite{ref_52} are also popular methods in handling concept drift. A more detailed analysis on diversity with the presence of different drift types is available from Minku et al.~\cite{ref_44}.


However, the concept drift literature hardly discusses seasonal concept drift. Consequently, the state-of-the-art concept drift classifiers are not capable of addressing seasonal concept drift. In this paper, we propose a novel classification algorithm that can be effective in high-dimensional stream classification with a high classification accuracy by addressing seasonal concept drift.

\section{Framework: Averaged One-Dependence Estimators}
\label{sec:framework}

AODE \cite{ref_3} is an improved version of NB that relaxes NB's attribute independence assumption. It has gained popularity since its introduction \cite{ref_28,ref_29,ref_46}.
%
%
AODE forms the base for building our novel classification model, SAODE. We choose AODE for this purpose due to the following reasons. 

\begin{enumerate}
\item AODE is a fast, updatable and accurate classifier.
\item Including a seasonal variable in a special way is easy with AODE. 
\end{enumerate}

In the following, we denote: $Y$ as the set of possible classes such that $Y = \langle y_1, y_2,...y_k \rangle$ where $k$ is the number of possible classes, $X$ as a sample of data such that $X = \langle x_1, x_2,..., x_n \rangle$ where $x_i$ is the value of the $i^{th}$ attribute and $n$ as the number of attributes in the dataset used for classification. 











In AODE, each attribute depends on the class and a parent attribute \cite{ref_3}. A set of attributes is chosen as parent attributes according to the frequency of their values and for each of the selected parent attributes, it constructs a separate One-Dependence Estimator (ODE). In each ODE, the attributes depend on their corresponding parent attribute and the class. Finally, the probability estimation for a class given a particular set of attribute values is calculated by averaging the probability estimations provided by the set of ODEs. 
AODE classifies an instance using the following equation \cite{ref_2}:

\begin{equation}
P(y,X) = \frac {\sum_{i:1 \leq i  \leq n  \land F(x_i) \geq m}P(y,x_i) \prod_{j:1}^{n}P(x_j|y,x_i)}
{|\{i:1 \leq i \leq n \land F(x_i) \geq m\}|}\label{eq4}
\end{equation}

Here, $F(x_i)$ is the frequency of $x_i$ over the dataset and $m$ is the minimum attribute value frequency over the dataset to be considered as a parent attribute.
As a classification model seeks the class that maximizes the resulting term, AODE ultimately chooses the class for a given instance that maximizes \cite{ref_2}:
\begin{equation}
\argmaxA_{y \in Y} \Bigg(\sum_{i:1 \leq i  \leq n  \land F(x_i) \geq m} P(y,x_i) \prod_{j:1}^{n}P(x_j|y,x_i)\Bigg)\label{eq5}
\end{equation}

\begin{figure}[t]
\includegraphics[width=\columnwidth]{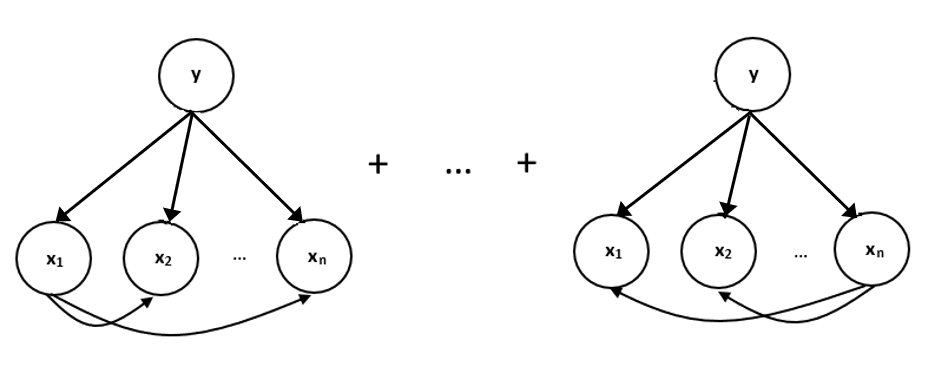}
\caption{A graphical representation of AODE. Let $X$ be an instance that needs to be classified such that $X = \langle x_1, x_2,..., x_n \rangle$ where $x_i$ is the value of the $i^{th}$ attribute, $n$ is the number of attributes in the dataset and $y$ is the instance label. AODE is a combination of ODEs and in each ODE, each attribute depends on $y$ and a parent attribute.} 
\label{AODE}
\end{figure}

Fig.~\ref{AODE} illustrates the main idea of AODE.

\section{The Proposed Classifier: Seasonal Averaged One-Dependence Estimators}
\label{sec:saode}

We seek to increase the accuracy of AODE by integrating it with time and seasonality factors. 
Therefore, as an extension to AODE, we make the time factor a super parent attribute of all other attributes in the dataset as follows:
%
%
\begin{equation}
P(y,X) \propto \frac {\splitdfrac{\sum_{i:1 \leq i  \leq n  \land F(x_i) \geq m} P(y,x_i,x_t)}{\prod_{j:1}^{n}P(x_j|y,x_i,x_t)}}
{|\{i:1 \leq i \leq n \land F(x_i) \geq m\}|}\label{eq6}
\end{equation}

Here, $x_t$ is the value of the time attribute and $n$ is the total number of attributes other than the time attribute. 
The accuracy of the classification model can be further improved by considering the relationship between the class and the time attribute. In this case, seasonality plays a major role as the frequencies of the classes depend on the time period that the classifying sample belongs to. This can be also considered as a weighting mechanism which gives more weight for a particular class according to the estimation of the posterior probability of the seasonal factor given the class. Therefore, the probability estimation of $y$ given $X$ can be calculated as follows:

\begin{equation}
P(y,X) = \dfrac {\splitdfrac{\sum_{i:1 \leq i  \leq n  \land F(x_i) \geq m} P(y)P(x_t|y)P(y,x_i,x_t)}{ \prod_{j:1}^{n}P(x_j|y,x_i,x_t)}}
{|\{i:1 \leq i \leq n \land F(x_i) \geq m\}|}\label{eq7}
\end{equation}

As the term $P(y)P(x_t|y)$ is common for all terms inside the sum, \eqref{eq7} can be further reduced as follows:

\begin{equation}
P(y,X) = \dfrac {\splitdfrac{P(y)P(x_t|y)\sum_{i:1 \leq i  \leq n  \land F(x_i) \geq m} P(y,x_i,x_t)}{ \prod_{j:1}^{n}P(x_j|y,x_i,x_t)}}
{|\{i:1 \leq i \leq n \land F(x_i) \geq m\}|}\label{eq8}
\end{equation}

A classification model seeks the class that maximizes the resulting term, and therefore SAODE selects the class for a given instance that maximizes:

\begin{multline}
\argmaxA_{y \in Y} \Bigg(P(y)P(x_t|y)\\\sum_{i:1 \leq i  \leq n  \land F(x_i) \geq m} P(y,x_i,x_t) \prod_{j:1}^{n}P(x_j|y,x_i,x_t)\Bigg)\label{eq9}
\end{multline}

\begin{figure}[t]
\includegraphics[width=\columnwidth]{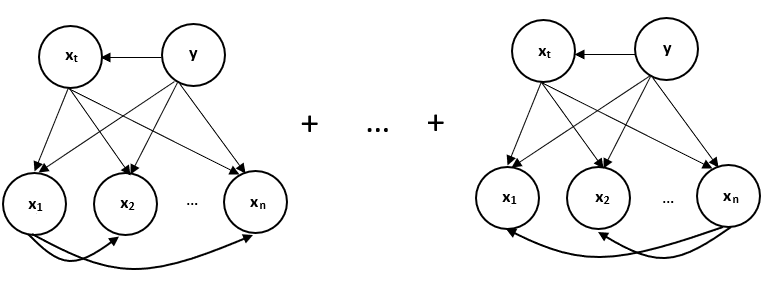}
\caption{A graphical representation of SAODE. Let $X$ be an instance that needs to be classified such that $X = \langle x_1, x_2,..., x_n,x_t \rangle$ where $x_i$ is the value of the $i^{th}$ attribute, $x_t$ is the value of the time attribute, $n$ is the number of attributes in the dataset other than the time attribute and $y$ is the instance label. SAODE is a combination of ODEs and in each ODE, each attribute depends on $y$, $x_t$ and one other parent attribute. SAODE considers the relationship between $y$ and $x_t$ as well.} 
\label{fig1}
\end{figure}

Fig.~\ref{fig1} illustrates the main idea of SAODE. Algorithm~\ref{alg1} shows the SAODE training/updating process and Algorithm~\ref{alg2} shows the SAODE classification process using the calculated frequencies of Algorithm~\ref{alg1}. 






\subsection{Relationship to A2DE}
A2DE is a special case of the Averaged n-Dependence Estimators (AnDE) model where $n=2$. In this case, each attribute depends on the class and all possible size two attribute sets in the dataset \cite{ref_4}. 


SAODE is related to the concept of A2DE as well. As in A2DE, it also constructs the classification model by establishing two parent attributes for each attribute in the dataset, namely the time attribute and one other attribute. But it does not consider all possible attribute pairs as parent attribute sets as in A2DE. 

Despite the close relationship to A2DE, SAODE has the following distinctive features which make it an attractive topic of study.

\begin{enumerate}
\item A2DE does not consider seasonality and time factors in its model construction. Also, it does not contain any weighting mechanism which gives more weight to a prediction estimation considering the relationship between the time and the class label.

\item A2DE is not feasible with large amounts of attributes as it considers all possible pairwise attribute combinations as parent attribute sets which eventually creates an ensemble of many models. Computational cost can therewith be considerably higher for A2DE than for SAODE. 
\end{enumerate}

\begin{algorithm}
\begin{algorithmic}[1]
\State \textbf{Initialization}
\State $count$, $f\_counts$, $c\_counts$, $t\_counts$ $\gets$ 0 
\State $a\_counts$, $av\_counts$ $\gets$ 0 
\State
\Procedure{train\_saode}{$X^*$, $n$} 
\For{$x$ in $X^*$}
\State $count$ $\gets$ $count$ + 1 
\State $c\_counts[y]$ $\gets$ $c\_counts[y]$ + 1
\If{$x_t$ is known}
\State $t\_counts[x_t]$ $\gets$ $t\_counts[x_t]$ + 1
\For{$i$ in 1 to $n$}
\If{$x_i$ is known}
\State $a\_counts[i]$ $\gets$ $a\_counts[i]$ + 1
\State $av\_counts[x_i]$ $\gets$ $av\_counts[x_i]$ + 1
\For{$j$ in 1 to $n$}
\If{$x_j$ is known}
\State $f\_counts[y, x_t, x_i, x_j]$ $\gets$ $f\_counts[y, x_t, x_i, x_j]$ + 1
\EndIf
\EndFor
\EndIf
\EndFor
\EndIf
\EndFor
\EndProcedure
\caption{SAODE training algorithm: Given the training set $X^*$ and the number of attributes $n$ other than the time attribute, the algorithm returns a 4-dimensional joint frequency table $f\_counts$, class frequency vector $c\_counts$, time-value frequency vector $t\_counts$, attribute frequency vector $a\_counts$, attribute-value frequency vector $av\_counts$ and item count $count$. The procedure $TRAIN\_SAODE$ is used to train SAODE. As a stream classifier, SAODE updates all frequency vectors using the procedure $TRAIN\_SAODE$ when a new set of instances $X^*$ is received with one or more examples.}
\label{alg1}
\end{algorithmic}
\end{algorithm}

    
\begin{algorithm}
\begin{algorithmic}[1]
\Procedure{classify\_saode}{$X$, $n$, $k$, $m$, $f\_counts$, $c\_counts$, $t\_counts$, $a\_counts$, $av\_counts$, $count$} 
\For{$y$ in 1 to $k$}
\State $prob[y]$ $\gets$ 0.0
\State $p\_count$ $\gets$ 0
\If{$x_t$ is known}
\For{$i$ in 1 to $n$}
\If{$x_i$ is known \AND $av\_counts[x_i] > m$}
\State $p\_count$ $\gets$ $p\_count$ + 1
\State $p$ $\gets$ $P(y, x\_i, x\_t)$
\For{$j$ in 1 to $n$}
\If{$x_j$ is known}
\State $p$ $\gets$ $p$ * $P(x\_j|y, x\_i, x\_t)$
\EndIf
\EndFor
\State $prob[y]$ $\gets$ $prob[y]$ + $p$
\EndIf
\EndFor
\State $prob[y]$ $\gets$ $prob[y]$ * $p(x\_t|y)$
\EndIf
\State $prob[y]$ $\gets$ $prob[y]$ * $p(y)$
\If {$p\_count$ $=$ 0}
\State $prob[y]$ $\gets$ $NB(X,y)$ \Comment{Use NB to classify $X$}
\Else
\State $prob[y]$ $\gets$ $prob[y]/p\_count$
\EndIf
\EndFor
\State Choose $c$ where $max(prob) = prob[c]$
\EndProcedure
\caption{SAODE classification algorithm: Given an instance $X$ that needs to be classified such that $X = \langle x_1, x_2,..., x_n,x_t \rangle$, the number of attributes $n$ other than the time attribute, the number of possible classes $k$, minimum attribute value frequency over the dataset to be considered as a parent attribute $m$ and the frequency vectors obtained during the training process: $f\_counts$, $c\_counts$, $t\_counts$, $a\_counts$, $av\_counts$ and $count$, the algorithm returns the index of the corresponding class of $X$.}
\label{alg2}
\end{algorithmic}
\end{algorithm}

\section{Experimental Methodology and Results}
\label{sec:exp}

In this section, we evaluate the proposed SAODE classifier using two real-world datasets. In particular, we evaluate SAODE against nine state-of-the-art stream and concept drift classification models: Hoeffding Tree \cite{ref_35}, Hoeffding Option Tree \cite{ref_34}, OzaBag \cite{ref_40}, OzaBoost \cite{ref_40}, Hoeffding Adaptive Tree \cite{ref_33}, Accuracy Updated Ensemble \cite{ref_36}, Leveraging Bagging \cite{ref_37}, NB and AODE\cite{ref_3} each with and without the consideration of the seasonal feature. We also compare SAODE with the variations of NB, Hoeffding Tree and AODE containing multiple classifiers: one for each considered season.

\subsection{Datasets}
We use the following two datasets related to text-mining, which is a real-world application with a high-dimensional feature space. 

\subsubsection{RCV1-v2 Dataset}
This dataset contains over 800,000 manually categorized news stories collected between August 1996 and August 1997. RCV1-v2 is a revised version of the RCV1 dataset, where data with errors have been removed \cite{ref_1}. Each news story in the RCV1-v2 dataset includes a set of words without any stop words and is related to one or more topics. The topic set used for labelling has four parent categories: \textit{Market}, \textit{Economic}, \textit{Government/Social} and \textit{Corporate/Industrial}. Therefore, the classification problem related to this dataset is mapping one or more parent level categories to each news story in the dataset. 
A seasonal feature (e.g. day of week) is required with each news story for the SAODE classification. The article date is used to calculate the seasonal feature. 


\subsubsection{NYT Dataset}
This dataset contains over 1.8 million articles published in the NYT newspaper between January 1, 1987 and June 19, 2007 \cite{ref_30}. Around 97.73\% of articles contain one or more online sections where the articles are placed on NYTimes.com. Therefore, the classification problem related to this dataset is mapping one or more online sections for articles in the dataset. The article date is used to calculate the seasonal feature similar to the RCV1-v2 dataset.

\subsection{Data Pre-Processing}
We first pre-process the data before applying the classification models. The following pre-processing techniques are applied to the datasets.

\subsubsection{Pre-processing of RCV1-v2 Dataset}

We use the most frequent words in the dataset to represent each news story. Choosing the best features for text classification seems beyond the scope of our work, as well as including more appropriate embeddings such as Word2Vec \cite{ref_49}. Rankings of classifiers are independent of text features as we use the same set of text features with all classifiers.
To identify the most frequent text features, first we identify the unique words belonging to each news story. The frequency of each unique word is calculated over the dataset and the top 2000 words containing the highest frequency are chosen for feature representation in the training data. Each feature indicates the presence or absence of a particular word in a news story.
We assume the word frequency is stable over the years in a real-world application.
A time feature is included in the dataset as the seasonal feature to indicate the day of the week. Finally, the training dataset used with SAODE contains 2001 features including the seasonal feature.
Each category combination is also mapped to a single class and therefore, SAODE can estimate the probability of each class at the testing time and map the highest possible class to each news story. 

\subsubsection{Pre-processing of NYT Dataset}
We identify the top four online sections: \textit{Arts}, \textit{Business}, \textit{Opinion} and \textit{New York and Region} that demonstrate the highest frequency over the full dataset. Therefore, the final dataset that is used for model building and evaluation contains 985,095 articles. The lead paragraphs of the articles are used to identify the text features after removing their stop words. The remaining pre-processing techniques include identifying the top text features, adding the seasonal feature and assigning classes to each article. They are similar to the pre-processing techniques we use with the RCV1-v2 dataset. Finally, the training dataset used with SAODE contains 2001 features including the seasonal feature, as in the RCV1-v2 dataset.

\begin{table}
\caption{AP, HL, MLA, MLFS and RMSE for all considered models for the  RCV1-v2 dataset with and without the presence of the seasonal feature. SAODE outperforms all other classifiers in all five evaluation metrics.}\label{tab1}
		\centering\fontsize{7}{8}\rm
		\begin{tabular}{rrrrrr}
			\toprule
			\cmidrule{2-6}
			& \ AP & HL    & MLA       & MLFS    & RMSE      \\\cmidrule{2-6}
			\addlinespace
			\multicolumn{6}{l}{\bf Classifiers without Seasonal Features} \\
			\addlinespace
			Accuracy Updated Ensemble & 48.1   & 0.176    &0.515   & 0.527    & 0.366  \\
			Hoeffding Tree & 74.8   & 0.095    & 0.831      & 0.859    & 0.296     \\
			Hoeffding Option Tree & 48.2   & 0.192    & 0.523      & 0.537    & 0.376     \\
			Hoeffding Adaptive Tree & 32.9    & 0.251      & 0.380    & 0.400 & 0.431     \\
			Leveraging Bagging& 58.5    & 0.171      & 0.675    & 0.708 &  0.353     \\
			OzaBag     & 55.1   & 0.163    & 0.603   & 0.621    & 0.352  \\
			OzaBoost & 47.8   & 0.265    & 0.583      & 0.629    & 0.383  \\
			NB     & 68.8   & 0.121    & 0.795   & 0.832    & 0.332  \\
			AODE  & 82.7   & 0.085  & 0.894  & \textbf{0.916}  & 0.294 \\
			\bottomrule
			\addlinespace
			\multicolumn{6}{l}{\bf Classifiers with Seasonal Features} \\
			\addlinespace
			Accuracy Updated Ensemble & 38.8   & 0.202    & 0.415   & 0.424    & 0.384  \\
			Hoeffding Tree & 68.5   & 0.118    & 0.762    & 0.789  & 0.322     \\
			Hoeffding Option Tree & 48.2   & 0.191    & 0.521      & 0.535    & 0.377     \\
			Hoeffding Adaptive Tree & 32.9    & 0.251    & 0.381      & 0.401    & 0.431    \\
			Leveraging Bagging& 59.3   & 0.168    & 0.684      & 0.718    & 0.352     \\
			OzaBag     & 55.1   & 0.171    & 0.615   & 0.639    & 0.356  \\
			OzaBoost & 41.2   & 0.308    & 0.533      & 0.589    & 0.402  \\
			NB     & 68.8   & 0.120    & 0.795   & 0.832    & 0.331  \\
			AODE & 82.8   & 0.085  & 0.894  & \textbf{0.916}  & 0.294 \\
			\bottomrule
			\addlinespace
			\multicolumn{6}{l}{\bf Multiple Classifiers, one per Season} \\
			\addlinespace
			Hoeffding Tree &  72.4   &  0.099  &  0.812   & 0.842   & 0.299 \\
			NB  & 70.1 & 0.103  & 0.802  & 0.837    & 0.302     \\
			AODE & 82.8   & 0.084  & 0.894  & \textbf{0.916}  & 0.294 \\
			\bottomrule
			\addlinespace
			\multicolumn{6}{l}{\bf Proposed Classifier} \\
			\addlinespace
			\textbf{SAODE} & \textbf{84.0}   & \textbf{0.083}  & \textbf{0.896}  & \textbf{0.916}  & \textbf{0.293} \\
			\bottomrule
		\end{tabular}\vspace*{0.1cm}
\end{table}

\begin{table}
\caption{AP, HL, MLA, MLFS and RMSE for all considered models for the NYT dataset with and without the presence of the seasonal feature. SAODE outperforms all other classifiers in all five evaluation metrics.}\label{tab2}
		\centering\fontsize{7}{8}\rm
		\begin{tabular}{rrrrrr}
			\toprule
			\cmidrule{2-6}
			& \ AP & HL    & MLA       & MLFS    & RMSE      \\\cmidrule{2-6}
			\addlinespace
			\multicolumn{6}{l}{\bf Classifiers without Seasonal Features} \\
			\addlinespace
			Accuracy Updated Ensemble   & 57.6  & 0.133  & 0.589 & 0.593 & 0.317\\
            Hoeffding Tree  & 80.0  & 0.092  & 0.813 & 0.818 & 0.279\\
            Hoeffding Option Tree   & 55.3  & 0.151  & 0.566 & 0.570 & 0.332\\
            Hoeffding Adaptive Tree & 38.1  & 0.203  & 0.408 & 0.417 & 0.396\\
            Leveraging Bagging  & 63.1  & 0.139  & 0.661 & 0.670 & 0.319\\
            OzaBag  & 57.8  & 0.140  & 0.591 & 0.596 & 0.319\\
            OzaBoost    & 56.7  & 0.138  & 0.581 & 0.586 & 0.316\\
            NB  & 73.7  & 0.119  & 0.761 & 0.768 & 0.316\\
            AODE  & 78.4  & 0.093  & 0.804 & 0.811 & 0.281\\
			\bottomrule
			\addlinespace
			\multicolumn{6}{l}{\bf Classifiers with Seasonal Features} \\
			\addlinespace
	        Accuracy Updated Ensemble  & 69.3  & 0.108  & 0.706 & 0.710 & 0.288\\
            Hoeffding Tree   & 81.7  & 0.085  & 0.827 & 0.830 & 0.270\\
            Hoeffding Option Tree   & 67.1  & 0.132  & 0.689 & 0.696 & 0.323\\
            Hoeffding Adaptive Tree   & 51.8  & 0.171  & 0.537 & 0.544 & 0.363\\
            Leveraging Bagging    & 64.1  & 0.129  & 0.663 & 0.670 & 0.309\\
            OzaBag   & 68.9  & 0.122  & 0.707 & 0.714 & 0.295\\
            OzaBoost     & 68.1  & 0.109  & 0.693 & 0.697 & 0.283\\
            NB  & 75.4  & 0.114  & 0.769 & 0.775 & 0.311\\
            AODE  & 80.2  & 0.086  & 0.815 & 0.819 & 0.273\\
			\bottomrule
			\addlinespace
			\multicolumn{6}{l}{\bf Multiple Classifiers, one per Season} \\
			\addlinespace
			Hoeffding Tree  & 81.6  & 0.085 &  0.826 & 0.829 & 0.272 \\
			NB  &  78.0  &  0.093  & 0.792  & 0.796 & 0.281 \\
            AODE & 82.4  & 0.084  & 0.835 & 0.838 & 0.271 \\
            \bottomrule
			\addlinespace
			\multicolumn{6}{l}{\bf Proposed Classifier} \\
			\addlinespace
			\textbf{SAODE}  & \textbf{83.5}  & \textbf{0.082}  & \textbf{0.845} & \textbf{0.848} & \textbf{0.265}\\
			\bottomrule
		\end{tabular}\vspace*{0.1cm}
\end{table}

\subsection{Evaluation}
We evaluate our model against nine state-of-the-art stream and concept drift classification models: Hoeffding Tree \cite{ref_35}, Hoeffding Option Tree \cite{ref_34}, OzaBag \cite{ref_40}, OzaBoost \cite{ref_40}, Hoeffding Adaptive Tree \cite{ref_33}, Accuracy Updated Ensemble \cite{ref_36}, Leveraging Bagging \cite{ref_37}, NB and AODE\cite{ref_3} each with and without the consideration of a seasonal feature. Additionally, we compare SAODE with three  models containing multiple classifiers: one for each considered season with the classification models: NB, Hoeffding Tree and AODE using \textit{prequential evaluation}. We implement SAODE using Weka \cite{ref_50} and use Weka and MOA (Massive Online Analysis) \cite{ref_51} built-in implementations to run baseline classifiers. All classifiers are run using their default parameters in Weka. All AODE variations including SAODE have one sole parameter, named $m$: the minimum attribute value frequency over the dataset to be considered as a parent attribute and we use $m=1$ that is the default value used in Weka.

Both classification problems are multi-label classifications. As we consider a small amount of labels, we can apply the single-label multi-class classifiers considered in this paper straightforwardly via a powerset approach. Due to this reason, we use multi-label metrics for the model evaluation.

We evaluate each model based on five metrics: Accuracy Percentage (AP), Hamming Loss (HL) \cite{ref_42}, Multi-Label Accuracy (MLA) \cite{ref_42}, Multi-Label $F_1$ Score (MLFS) \cite{ref_42} and Root Mean Squared Error (RMSE) \cite{ref_43}. They are discussed in the following.

\begin{figure*}[t]
\includegraphics[width=\textwidth]{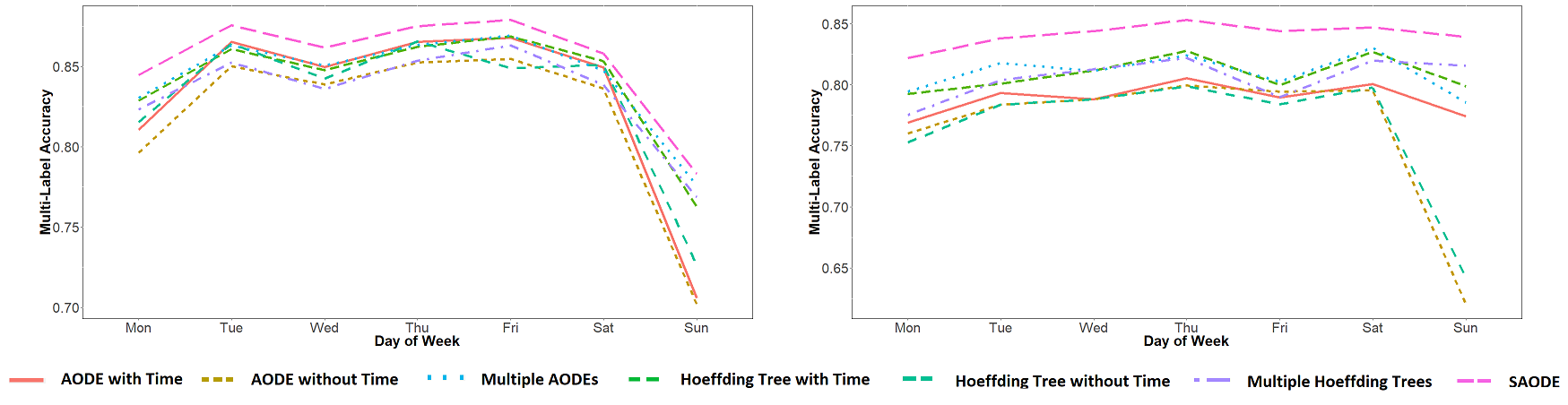}
\caption{Plot of MLAs calculated over day of the week for seven high performing classifiers: AODE with Time, AODE without Time, Hoeffding Tree with Time, Hoeffding Tree without Time, Multiple AODE Classifiers, Multiple Hoeffding Trees and SAODE. Left: MLAs calculated for the NYT dataset. Right: MLAs calculated for the New York and Region section of the NYT dataset.} 
\label{fig9}
\end{figure*}


Let $D$ refer to the set of input documents, $L$ is the set of possible labels, $N$ is the total number of instances in the test set, $N_c$ is the number of correctly labelled instances in the test set, $\lambda_{d,l}$ is the probability given by a classifier that document $d$ has label $l$, $H_d$ is the estimated set of labels for document $d$ and $Y_d$ is the actual set of labels for document $d$. For any Boolean expression $b$, $[[b]]$ returns $1$ if $b$ is true, otherwise, it returns $0$.

AP is the proportion of correctly classified instances of the test set as a percentage:

\begin{equation}
AP = \frac{N_c}{N} \times 100 \%\label{eq10}
\end{equation}

HL \cite{ref_42}: the average proportion of members of L that are incorrectly predicted is defined as

\begin{equation}
HL = \frac{1}{|D|}\sum_{d \in D}\frac{|H_d \triangle Y_d|}{|L|}\label{eq11}
\end{equation}

Here, $\triangle$ is the symmetric difference between sets.

MLA \cite{ref_42}: the average fraction of labels that are correctly predicted is defined as

\begin{equation}
MLA = \frac{1}{|D|}\sum_{d \in D}\frac{|H_d \cap Y_d|}{|H_d \cup Y_d|}\label{eq12}
\end{equation}

This is the Jaccard Index of the predicted and actual label sets.

MLFS~\cite{ref_42}: the harmonic mean between the precision and recall is defined as

\begin{equation}
MLFS = \frac{1}{|D|}\sum_{d \in D}\frac{2|H_d \cap Y_d|}{|H_d| + |Y_d|}\label{eq13}
\end{equation}

RMSE \cite{ref_43}: the root mean square of the difference between the predicted probability and the actual value for each label is defined as

\begin{equation}
RMSE = \sqrt{\frac{\sum_{d \in D}\sum_{l \in L}(\lambda_{d,l} - [[l \in Y_d]])^2}{|D||L|}}\label{eq14}
\end{equation}

Tables~\ref{tab1} and \ref{tab2} report results across the five metrics for all considered models for the RCV1-v2 dataset and the NYT dataset, respectively. 

From the tables we can see that the proposed classifier, SAODE, outperforms all other considered state-of-the-art classification models for both datasets, consistently across all error measures. We also performed pairwise tests for statistical significance using a Wilcoxon test~\cite{ref_56} with a Bonferroni correction for all methods against our method and all results were highly significant ($p$-value$\ < 10^{-16}$).

AODE provides the highest accuracy of 0.894 for the RCV1-v2 dataset in classifying data without seasonal features and SAODE improves this slightly to 0.896. The Hoeffding Tree provides the highest accuracy of 0.813 in classifying data without seasonal features for the NYT dataset and SAODE improves this to 0.845. Furthermore, the accuracy of the majority of classifiers increases when using a seasonal feature, for the NYT dataset. But the accuracies of Accuracy Updated Ensemble, Hoeffding Tree, Hoeffding Option Tree and OzaBoost decrease when using a seasonal feature for the RCV1-v2 dataset.

However, just including an additional seasonal feature does not considerably improve the accuracy of the models, due to the large number of features used in this classification task. 

SAODE gives the seasonal feature a special role by making it a super parent of all other features and is therewith able to achieve a higher accuracy. It outperforms all nine considered state-of-the-art classification models.

We further investigate the classification accuracy of SAODE for separate days of the week. The left-hand side of Fig.~\ref{fig9} illustrates this concept by plotting the calculated MLAs of SAODE and other six high performing classifiers: AODE with time, AODE without time, Hoeffding Tree with time, Hoeffding Tree without time, multiple AODE classifiers and multiple Hoeffding Trees for each day of the week in the NYT dataset. The accuracy slightly increases for each day when using a seasonal feature. The variations of the Hoeffding Tree also perform better compared to the basic versions of AODE: AODE with time and AODE without time. The variation of AODE containing multiple classifiers, namely one per each season, further improves the classification accuracy of each day, but SAODE outperforms all of these high performing classifiers for all days of the week in classification accuracy.
%

\begin{figure}[t]
\includegraphics[width=\columnwidth]{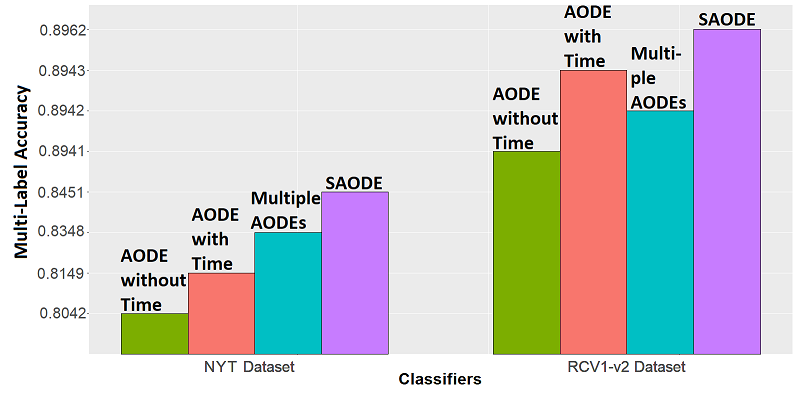}
\caption{Plot of MLAs calculated with the RCV1-v2 dataset and the NYT dataset for the variations of AODE: AODE without Time, AODE with Time and Multiple AODEs for each seasonality and for our classifier: SAODE.}
\label{fig13}
\end{figure}

The right-hand side of Fig.~\ref{fig9} illustrates another pattern we observe in classification accuracy. It shows the calculated MLAs for the New York and Region section articles in the NYT dataset for all days of the week for the same set of classifiers. It further shows SAODE outperforms all high performing classifiers consistently for each day of the week in classification accuracy. Additionally, it shows how SAODE here is able to maintain high classification accuracy on Sundays, compared to other classifiers whose accuracy degrades on this particular day of the week. This illustrates the ability of SAODE to address the seasonal concept drift of this dataset.
We investigate the performance of AODE variations in Fig.~\ref{fig13}. The classification accuracy slightly increases when using a seasonal feature for both datasets. The variation of AODE including multiple classifiers for each considered season further improves the accuracy of AODE with the NYT dataset and decreases the accuracy of AODE with the RCV1-v2 dataset. SAODE uses a seasonal feature as a super parent of the other features, and it considers the relationship between the class label and the seasonality whereas the traditional model does not consider this relationship. As a result of this, SAODE outperforms all AODE variations in classification accuracy for both datasets.


\section{Conclusion}
\label{sec:conclusion}

Seasonal concept drift is a phenomenon commonly observed that needs to be exclusively investigated. In this paper, we have proposed a novel classification model, SAODE that is designed for high-dimensional stream classification with the consideration of seasonal concept drift. To address the seasonal concept drift, SAODE builds on the AODE classifier by combining it with a special seasonal feature. It then considers the relationship between the seasonal feature and the class label as well as the seasonal feature and every other feature of the dataset. We have tested SAODE with two large real-world datasets and it is able to consistently outperform nine state-of-the-art stream and concept drift classification models across different error measures where the results are statistically significant.

The success of this approach encourages as future work to build a classification model for more complex seasonalities, and to use multiple seasonal features to classify data that exhibit multiple seasonalities (e.g., daily, weekly, and quarterly). Next we will look at extending SAODE to work with continuous attributes, for instance by using the incremental discretizers that have been developed for stream data. 


\bibliographystyle{IEEEtran}
\bibliography{references}

\end{document}